\documentclass{article}

\usepackage{arxiv}

\usepackage[utf8]{inputenc} 
\usepackage[T1]{fontenc}    
\usepackage{hyperref}       
\usepackage{url}            
\usepackage{booktabs}       
\usepackage{amsfonts}       
\usepackage{nicefrac}       
\usepackage{microtype}      
\usepackage{lipsum}
\usepackage{graphicx}
\usepackage{amsmath} 
\usepackage{threeparttable}
\usepackage{makecell}
\usepackage{booktabs}
\usepackage{longtable}
\usepackage{multirow}
\usepackage{booktabs}
\usepackage{threeparttable}
\usepackage{multirow}
\usepackage{siunitx}
\usepackage{array}
\usepackage{tikz}
\usepackage{pgfplots}
\pgfplotsset{compat=1.18}
\graphicspath{ {./images/} }
\usepackage{pgfplots}
\usepgfplotslibrary{groupplots}
\pgfplotsset{compat=1.18}
\usepackage{booktabs,tabularx,ragged2e,xurl} 
\newcolumntype{P}[1]{>{\RaggedRight\arraybackslash}p{#1}}
\newcolumntype{Y}{>{\RaggedRight\arraybackslash}X}
\newcommand{\qcode}[1]{\path{#1}} 
\usepackage{tabularx}
\usepackage{booktabs}
\usepackage{dblfloatfix} 
\usepackage{placeins}    
\usepackage{float}
\usepackage{booktabs}
\usepackage{tabularx}
\usepackage{ltablex}   
\usepackage{caption}
\usepackage{array}
\keepXColumns

\title{Which Quantization Should I Use? A Unified Evaluation of llama.cpp Quantization on Llama-3.1-8B-Instruct}

\author{
 Uygar Kurt \\
}

\begin{document}
\maketitle
\begin{abstract}
Quantization is a practical technique for making large language models easier to deploy by reducing the precision used to store and operate on model weights. This can lower memory use and improve runtime feasibility on constrained hardware, which is especially relevant for users running models locally. Quantization in \texttt{llama.cpp} enables large language models to run on commodity hardware, but available formats are often evaluated inconsistently, making it hard to choose among schemes. We present a unified empirical study of the \texttt{llama.cpp} quantization on a single modern model, Llama-3.1-8B-Instruct (FP16, GGUF), covering 3–8 bit K-quant and legacy formats. We evaluate downstream task performance across standard reasoning, knowledge, instruction-following, and truthfulness benchmarks, and also measure perplexity and CPU throughput (prefill/decoding) alongside model size, compression, and quantization time. Ultimately, this work is a practical guide for choosing a \texttt{llama.cpp} quantization scheme, helping readers make informed, context-aware decisions for their intended use and resource budget.
\end{abstract}

\section{Introduction}
Large language models (LLMs) have rapidly become a core building block of modern AI systems. Yet, their computational and memory requirements remain a significant barrier to deployment on non-datacenter-scale hardware. A dense 8–70 B parameter model in full-precision FP16 commonly requires tens to hundreds of gigabytes of memory for inference, placing it beyond the reach of commodity CPUs, consumer GPUs, and edge devices. Quantization replaces high-precision floating-point weights and activations with low-precision integer or custom formats. It has therefore become a central technique for making LLMs practical in resource-constrained settings. Prior work has demonstrated that 8-bit and even 4-bit representations can preserve most of the predictive performance of full-precision transformers while reducing memory and bandwidth costs by factors of 2 to 4 \cite{yao2022zeroquant}\cite{frantar2022gptq}\cite{dettmers2023spqr}, enabling the deployment of models with tens or hundreds of billions of parameters on a single accelerator.

The research community has developed a rich family of quantization methods for LLMs, including mixed-precision matrix multiplication (LLM.int8()) \cite{dettmers2022gpt3}, post-training weight-only methods such as GPTQ \cite{frantar2022gptq} and AWQ \cite{lin2024awq}, and fine-tuning approaches such as QLoRA \cite{dettmers2023qlora}. These methods are often evaluated in PyTorch-based pipelines using a small number of base models, and a large volume of recent work has begun to systematically compare their behavior across benchmarks and microscaling formats \cite{gong2024llmc}\cite{huang2024empirical}\cite{zheng2025empirical}\cite{lee2024exploring}\cite{zhao2025quantitative}\cite{liu2025quantization}. 

\texttt{llama.cpp} \footnote{https://github.com/ggml-org/llama.cpp} is a community-driven, highly optimized C/C++ inference engine targeting CPUs and lightweight GPUs. The primary goal of \texttt{llama.cpp} is to enable "LLM inference with minimal setup and state-of-the-art performance on a wide range of hardware," and it is widely adopted for running quantized open-weight models locally. To support this goal, \texttt{llama.cpp} implements a large set of weight-only quantization formats in the GGUF file format, including the original schemes ($\texttt{Q4\_0}$, $\texttt{Q4\_1}$, $\texttt{Q5\_0}$, $\texttt{Q5\_1}$, $\texttt{Q8\_0}$), which divides the weight matrices into groups, and the more recent "K-quant" family ($\texttt{Q2\_K}$, $\texttt{Q3\_K*}$, $\texttt{Q4\_K*}$, $\texttt{Q5\_K*}$, $\texttt{Q6\_K}$), which use superblocks and additional tricks to improve quality at a given model size. These formats have primarily been designed and iterated in an engineering-driven, community-centric workflow. New quantizers are introduced and tuned via GitHub issues, pull requests, and third-party model conversions, rather than through formal algorithmic descriptions and peer-reviewed evaluation. As a result, practical guidance on "which GGUF quantization to use" is dominated by anecdotal advice and perplexity measurements, with no unified experimental framework for comparison.

This situation poses a significant challenge. Users of \texttt{llama.cpp} must choose among a dozen or more quantization configurations that trade off model size, inference speed, and quality in non-obvious ways. Yet existing evaluations typically (i) focus on language-model perplexity on Wikitext-2 or similar corpora, (ii) mix results across heterogeneous base models and instruction-tuned variants, and (iii) rarely report downstream task performance in a unified way. Consequently, it is difficult to answer even basic questions such as "How much quality do I lose by moving from $\texttt{Q8\_0}$ to $\texttt{Q4\_K\_M}$ on a modern Llama-3-class model?" or "Do 3-bit K-quants remain usable on downstream tasks?" in a principled, model-specific manner.

In this paper, we address this gap by presenting a comprehensive empirical study of the \texttt{llama.cpp} quantization methods using widely available Llama-3.1-8B-Instruct \cite{grattafiori2024llama}. We quantify the trade-offs among model quality, memory footprint, and quantization cost across a broad set of GGUF formats, ranging from 3-bit to 8-bit weights. Concretely, we:
\begin{itemize}
    \item Evaluate 13 quantization configurations ($\texttt{Q3\_K*}$, $\texttt{Q4\_0/1}$, $\texttt{Q4\_K*}$, $\texttt{Q5\_0/1}$, $\texttt{Q5\_K*}$, $\texttt{Q6\_K}$, $\texttt{Q8\_0}$) alongside a full-precision FP16 baseline, all derived from the same Llama-3.1-8B-Instruct checkpoint and quantized with the official \texttt{llama.cpp} tooling.
	\item Measure downstream performance on a suite of established benchmarks that probe reasoning, world knowledge, and instruction-following: GSM8K \cite{cobbe2021training}, HellaSwag \cite{zellers2019hellaswag}, IFEval \cite{zhou2023instruction}, MMLU \cite{hendrycks2020measuring}, and TruthfulQA \cite{lin2022truthfulqa}. These scores complement perplexity on Wikitext-2 \cite{merity2017regularizing}, providing a richer view of how quantization affects actual task behavior.
    \item We conduct a thorough analysis of the size reduction, benchmark score reduction, and perplexity increase across different quantization schemes, on which strategy is best to use in which situation.
	\item Characterize efficiency trade-offs by reporting model size, compression ratio, and wall-clock quantization time on a dual-socket Intel Xeon Platinum 8488C system with AVX-512/BF16 support, using a common GGUF FP16 input model.
    \item We evaluate CPU inference throughput (prefill and decoding) for all quantization configurations, highlighting the impact of quantization choices on real-world inference performance.
    \item We open-sourced all quantized models generated in this paper on HuggingFace for the community to use and replicate our results \footnote{https://huggingface.co/uygarkurt/Llama-3.1-8B-Instruct-GGUF}.
\end{itemize}

By grounding our analysis in a single, strong, widely deployed model and a consistent evaluation pipeline, we aim to consolidate fragmented knowledge about the \texttt{llama.cpp} quantization into quantitative, reproducible evidence. Beyond providing practical recommendations for choosing GGUF quantization schemes for Llama-3.1-8B-Instruct, our study offers a template for future work on other architectures and backends, and highlights where the community-driven formats of \texttt{llama.cpp} aligns with, or diverges from, the behavior predicted by more general LLM quantization theory.

\section{Background \& Related Work}
\subsection{Post Training Quantization of Large Language Models}
Modern large language models (LLMs) are usually trained and distributed in floating-point (e.g., FP16/BF16/FP32). The \texttt{llama.cpp} toolchain targets efficient local inference by applying post-training quantization (PTQ) to the model weights and storing them in GGUF/GGML quantized tensor formats (e.g., $\texttt{Q4\_0}$, $\texttt{Q4\_1}$, $\texttt{Q5\_0}$, $\texttt{Q8\_0}$, $\texttt{Q*\_K}$, $\texttt{IQ*}$). At runtime, the model loads these quantized weight tensors—often via memory mapping—and executes optimized kernels that operate directly on the packed low-bit representations. Activations are generally maintained in floating-point precision (FP16/FP32/BF16 depending on the backend and build), although some quantized kernels temporarily quantize activations (e.g., to \texttt{Q8\_0}) within individual operations.

In \texttt{llama.cpp}, the GGML/GGUF formats use block-wise uniform quantization. They pack weights into quantized integer values and store a small amount of per-block metadata (a scale, and sometimes an offset) needed to map between real values and quantized integer values. Concretely, for a real scalar w and an integer code range $[q_{\min}, q_{\max}]$ with $2^b$ levels, an affine quantizer has the form
\begin{equation}
q(w) = \operatorname{clip}\!\Big(\big\lfloor \tfrac{w}{s}\big\rceil + z,\; q_{\min},\; q_{\max}\Big),
\qquad
\hat{w} = s\,(q(w)-z),
\label{eq:affine-quant}
\end{equation}
with scale $s>0$ and integer zero-point $z$. In GGML-style weight quantization, \eqref{eq:affine-quant} is instantiated per block rather than per tensor. For a block vector $v\in\mathbb{R}^K$ (typically $K=32$ for the standard formats), one stores block metadata $(s,z)$ (or an equivalent parameterization) and packed quantized integer values $q_i$ for each entry $v_i$.

This view makes the connection between named GGML quantization types and affine quantizers explicit. For example, $\texttt{Q4\_0}$ is a symmetric, block-wise 4-bit quantizer in which each block of $K=32$ floats stores a single scale and 4-bit signed quantized integer values that represent integers in $[-8,7]$ (packed as nibbles); dequantization is equivalent to a symmetric affine quantizer with $z=0$ and $\hat{v}_i = s\,q_i$. By contrast, $\texttt{Q4\_1}$ is an affine (asymmetric) block quantizer that stores both a block scale and a block offset (often described as a stored minimum), so that dequantization takes the form
\begin{equation}
\hat{v}_i = s\,q_i + m,
\qquad q_i \in \{0,1,\dots,15\},
\label{eq:q41}
\end{equation}
which corresponds to \eqref{eq:affine-quant} with a nonzero effective offset. The same block-wise pattern extends to other basic formats (\texttt{Q5\_0}, \texttt{Q5\_1}, \texttt{Q8\_0}), differing mainly in $b$ and in whether an offset/minimum is stored.

A useful first-order characterization of the accuracy--bit-width trade-off comes from the standard high-resolution approximation: away from clipping, the scalar quantization error $e=w-\hat{w}$ is modeled as uniform on $\left[-\tfrac{\Delta}{2},\tfrac{\Delta}{2}\right]$ with step size $\Delta$. For a symmetric $b$-bit uniform quantizer over $[-a,a]$ with $a=\max |w|$, one has $\Delta \approx \tfrac{2a}{2^b-1}$, giving
\begin{equation}
\mathbb{E}[e]\approx 0,
\qquad
\operatorname{Var}(e)\approx \frac{\Delta^2}{12}=\mathcal{O}(2^{-2b}),
\label{eq:var}
\end{equation}
so that increasing $b$ by one bit typically reduces the modeled variance by a factor of $\approx 4$ (subject to deviations from the assumptions due to clipping, heavy tails, and non-uniform error weighting). In \texttt{llama.cpp}, block-wise quantization changes the relevant $a$ and $\Delta$ from a global tensor range to a local block range, which is precisely why finer grouping (smaller blocks / more scales) often improves quality at a fixed nominal bit-width.

In the PTQ taxonomy, \texttt{llama.cpp} is primarily employs weight-only quantization. It quantizes weights offline into GGUF for inference, rather than storing quantized activations.

Quant types also differ by grouping/metadata. Standard formats quantize blocks of $K=32$ weights with a per-block scale (and sometimes an offset/min), whereas the \texttt{\_K} family uses $256$-weight super-blocks split into smaller groups with additional (often quantized) scale/min metadata to improve low-bit fidelity, i.e., a hierarchical affine quantizer.

\subsection{llama.cpp and Its Quantization Schemes}
\label{sec:llamacpp-quants}
\begin{table}[t]
    \centering
    \begin{tabular}{lcl}
        \toprule
        \textbf{Scheme} & \textbf{Bits (approx.)} & \textbf{Brief description / intuition} \\
        \midrule
        Q3\_K\_S   & $\sim 3$ & K-block 3-bit, ``small'' variant, emphasizes compression \\
        Q3\_K\_M   & $\sim 3$ & K-block 3-bit, ``medium'' trade-off \\
        Q3\_K\_L   & $\sim 3$ & K-block 3-bit, ``large'' / quality-oriented \\
        \midrule
        Q4\_0      & $\sim 4$ & Older 4-bit scheme, simple, widely available \\
        Q4\_1      & $\sim 4$ & 4-bit with different scaling design \\
        Q4\_K\_S/M & $\sim 4$ & Newer K-block 4-bit, tuned for speed / quality \\
        \midrule
        Q5\_0/1    & $\sim 5$ & 5-bit schemes, near-FP16 quality in many tasks \\
        Q5\_K\_*   & $\sim 5$ & K-block 5-bit, more modern variants \\
        \midrule
        Q6\_K      & $\sim 6$ & High-quality, near-FP16 \\
        Q8\_0      & $\sim 8$ & Very close to FP16, reference-ish baseline \\
        \bottomrule
    \end{tabular}
    \caption{Overview of the \texttt{llama.cpp} GGUF quantization schemes considered in this work, with approximate effective bit-widths and qualitative intent.}
    \label{tab:llamacpp-schemes}
\end{table}

\texttt{llama.cpp} is a C/C++ inference engine designed to run LLaMA and compatible transformer models with minimal dependencies and high efficiency across CPUs and GPUs. It relies on the GGML tensor library and the GGUF model format, which stores model weights (optionally quantized) alongside metadata for memory-mapped, zero-copy loading. Within this ecosystem, quantization is a first-class feature. Most deployment scenarios rely on quantized GGUF models rather than full-precision checkpoints.

The project supports a rich set of quantization formats, implemented as custom GGML tensor types. At a high level, these fall into two main families relevant to our study. Standard block-wise integer formats ($\texttt{Q4\_0}$, $\texttt{Q4\_1}$, $\texttt{Q5\_0}$, $\texttt{Q5\_1}$, $\texttt{Q8\_0}$) and the more recent K-quant formats ($\texttt{Q2\_K}$, $\texttt{Q3\_K\_S/M/L}$, $\texttt{Q4\_K\_S/M}$, $\texttt{Q5\_K\_S/M}$, $\texttt{Q6\_K}$).
All are weight-only and operate on fixed-size blocks or super-blocks, leaving
selected tensors (e.g.\ layer norms, embeddings) at higher precision. For convenience, these quantization types are summarized in Table \ref{tab:llamacpp-schemes}.

\paragraph{Standard integer formats.}
The classic or the legacy formats in \texttt{llama.cpp} are:
\begin{itemize}
  \item \textbf{Q4\_0, Q4\_1 (4-bit).}
    $\texttt{Q4\_0}$ is symmetric (no zero-point), while $\texttt{Q4\_1}$ adds a per-block offset/zero-point (affine quantization), which can better accommodate asymmetric weight distributions.

  \item \textbf{Q5\_0, Q5\_1 (5-bit).}
    The 5-bit analogs of $\texttt{Q4\_0/Q4\_1}$: symmetric (\_0) versus affine (\_1). The larger integer grid (e.g., typically spanning $\{-16,\dots,15\}$) can reduce quantization error at a modest memory increase.

  \item \textbf{Q8\_0 (8-bit).}
    Symmetric 8-bit quantization, which is commonly used as a high-fidelity baseline in GGUF pipelines, trades higher storage for accuracy closer to FP16 in many settings.
\end{itemize}

\paragraph{K-quant formats.}
The K-quant family introduces a hierarchical super-block structure that provides additional flexibility in allocating bits and metadata within a tensor. A typical K-quant format partitions weights into super-blocks of 256 values and stores multiple (often quantized) scale/min parameters per sub-block. Some variants also use mixed precision across sub-blocks. This leads to a non-integer effective bits-per-weight (bpw), e.g., $\approx 3.44$ bpw for \texttt{Q3\_K\_S} and $\approx 4.5$ bpw for \texttt{Q4\_K\_M}. The suffixes \texttt{\_S}, \texttt{\_M}, and \texttt{\_L} denote small/medium/large variants that trade off compression and fidelity. Within this family, we focus on:

\begin{itemize}
  \item \textbf{3-bit: Q3\_K\_S, Q3\_K\_M, Q3\_K\_L.}
    Variants with effective bit-widths of roughly $\approx 3.4$, $3.5$, and $3.6$ bits. \texttt{S} prioritizes smaller metadata and faster kernels, \texttt{M} balances speed/quality, and \texttt{L} allocates more information to improve reconstruction fidelity.

  \item \textbf{4-bit: Q4\_K\_S, Q4\_K\_M.}
    Effective bit-widths of roughly $\approx 4.2$ and $\approx 4.5$ bits; typically higher-fidelity than $\texttt{Q4\_0/Q4\_1}$ at similar (or slightly higher) size.

  \item \textbf{5-bit: Q5\_K\_S, Q5\_K\_M.}
    Effective bit-widths of roughly $\approx 5.2$ and $\approx 5.5$ bits, offering a further quality/size trade-off.

  \item \textbf{6-bit baseline: Q6\_K.}
    A high-fidelity K-quant with effective bit-width around $\approx 6.5$ bits, often used when accuracy is prioritized while still reducing memory versus FP16.
\end{itemize}

Crucially, these schemes are largely community-driven. New formats, effective bit estimates, and recommended usage patterns have emerged through GitHub issues, pull requests, and informal benchmark reports. While recent documentation has begun to systematize this knowledge, there remains no unified, model-specific evaluation that compares these formats on a modern LLM using a common pipeline.

\subsection{Existing Quantization Benchmarks}
\label{sec:existing-benchmarks}
A growing body of work examines how quantization impacts LLM quality and deployment cost, typically reporting trade-offs among perplexity, downstream accuracy, latency, and memory footprint. Much of this literature focuses on post-training quantization (PTQ) and related recipes such as LLM.int8(), GPTQ, AWQ, SpQR, SmoothQuant \cite{xiao2023smoothquant}, Quip\# \cite{tseng2024quip}, and QLoRA, and is evaluated across multiple model families (GPT\cite{brown2020language}, OPT\cite{zhang2022opt}, LLaMA\cite{touvron2023llama}\cite{touvron2023llama2}\cite{grattafiori2024llama}, Qwen \cite{yang2025qwen3}) and heterogeneous task suites. More recent empirical work has also expanded coverage to newer instruction-tuned and multimodal-capable backbones, including targeted studies of LLaMA~3 quantization behavior across bit-widths and methods \cite{huang2024empirical}, as well as stack- and model-specific analyses for newly released families such as Qwen3 \cite{zheng2025empirical}.

Recent studies also move toward more standardized, multi-method comparisons. Jin et al.\cite{jin2024comprehensive} provide a broad evaluation of quantization strategies across multiple model scales, emphasizing efficiency--quality trade-offs and discussing when perplexity can serve as a proxy for downstream behavior. Complementary benchmarking efforts, including the qLLM-eval framework \cite{li2024evaluating} and related evaluations of quantized LLMs, compare multiple PTQ approaches (e.g., AWQ and SmoothQuant) across tasks spanning language modeling, classification, and question answering \cite{li2024evaluating}. LLMC \cite{gong2024llmc} further pushes in this direction by offering a unified compression toolkit intended to enable fairer comparisons across a wide range of quantization configurations, numeric formats, and model types (including vision-language settings). At the same time, the community has increasingly emphasized behavioral coverage beyond easy accuracy proxies: large-scale evaluations across instruction-following and hallucination-style benchmarks highlight that quantization effects can vary substantially with task difficulty and model size \cite{lee2024exploring}. Other reports are more domain- or stack-specific, such as cross-lingual analyses for LLaMA-family models or industrial evaluations that focus on particular formats (e.g., FP8) and hardware/software ecosystems \cite{iakovenkocomparing}. In parallel, dedicated studies on reasoning-focused models and benchmarks show that aggressive low-bit settings can be especially risky for multi-step reasoning, and that conclusions can depend on whether weights, activations, and/or KV cache are quantized \cite{liu2025quantization}.

Despite this progress, the existing evidence is often less actionable for \texttt{llama.cpp} style deployment. First, many evaluations remain perplexity-centric (often on corpora such as WikiText-2 or C4) and include only limited downstream coverage; while perplexity is informative, it does not fully characterize instruction-following, reasoning, or safety-relevant behavior that dominates real-world use of instruction-tuned models. Second, results are frequently aggregated across different base models and task suites, which confounds direct, format-to-format comparisons and makes it difficult to isolate the interaction between bit-width, quantization scheme, and a single modern architecture (e.g., Llama~3.1--8B). Third, most academic benchmarks operate in PyTorch/JAX with generic quantization operators, and to our knowledge do not systematically evaluate the GGUF-specific quantization formats as implemented in \texttt{llama.cpp} (e.g., $\texttt{Q3\_K\_S/M/L}$, $\texttt{Q4\_K\_S/M}$, $\texttt{Q5\_K\_S/M}$, $\texttt{Q6\_K}$, and $\texttt{Q4\_0/Q4\_1/Q5\_0/Q5\_1/Q8\_0}$), despite their widespread practical use. Even model-specific deployment-oriented studies \cite{zhao2025quantitative} typically compare quantization at the level of bit-width or strategy, rather than at the level of concrete GGUF formats, and therefore do not directly inform GGUF format selection in \texttt{llama.cpp} pipelines.

Our study addresses this gap by fixing a single widely deployed model (Llama-3.1-8B Instruct), quantizing it with the official \texttt{llama.cpp} toolchain across a broad set of GGUF formats, and evaluating all variants under a unified pipeline. We report both perplexity and a diverse set of downstream benchmarks (e.g., GSM8K, HellaSwag, IFEval, MMLU, TruthfulQA), while controlling the base model, evaluation harness, and hardware environment to enable an apples-to-apples comparison of \texttt{llama.cpp} quantization schemes that complement prior method-centric studies and newer multi-task evaluations.

\section{Experimental Setup}
\label{sec:setup}

\subsection{Base Model \& Environment}
\label{sec:setup:base}
All experiments use \texttt{Llama-3.1-8B-Instruct} converted to GGUF and evaluated in \texttt{F16} as the reference (original) model. For all evaluations, prompts were formatted using the default Llama-3.1-8B-Instruct chat template. The resulting input GGUF file size is $15317.02$~MiB ($\approx 14.96$~GiB). Quantization and evaluation were performed on a dual-socket CPU server with two Intel Xeon Platinum 8488C processors, totaling $96$ physical cores ($192$ threads), with AVX-512 and BF16 support enabled. We used \texttt{llama.cpp} (commit b7600) \footnote{\url{https://github.com/ggml-org/llama.cpp/tree/b7600}} for GGUF quantization and perplexity/throughput measurements, and the EleutherAI evaluation harness lm\_eval (v0.4.9.2) \footnote{\url{https://github.com/EleutherAI/lm-evaluation-harness/tree/v0.4.9.2}} for downstream benchmark scoring. We used the default evaluation harness and \texttt{llama.cpp} configurations for downstream benchmark scoring and perplexity scoring.

\subsection{Quantization Configurations}
\label{sec:setup:quant}
We evaluate community-standard \texttt{llama.cpp} GGUF quantization schemes spanning nominal 3--8 bit weight formats:
\(\{\texttt{Q3\_K\_S}, \texttt{Q3\_K\_M}, \texttt{Q3\_K\_L}\}\),
\(\{\texttt{Q4\_0}, \texttt{Q4\_1}, \texttt{Q4\_K\_S}, \texttt{Q4\_K\_M}\}\),
\(\{\texttt{Q5\_0}, \texttt{Q5\_1}, \texttt{Q5\_K\_S}, \texttt{Q5\_K\_M}\}\),
\(\{\texttt{Q6\_K}\}\), and \(\{\texttt{Q8\_0}\}\).
All quantized models are produced from the same FP16 GGUF reference input using the official \texttt{llama.cpp}
quantization utility (e.g., \texttt{llama-quantize}/\texttt{quantize}), with the scheme identifier as the only varying
parameter. A canonical command pattern is:
\begin{verbatim}
./llama-quantize <f16.gguf> <out.gguf> <SCHEME>
\end{verbatim}
We measure wall-clock quantization time on the same machine used for inference.

Let \(S_{\text{F16}}\) denote the FP16 input model size and \(S_q\) the resulting quantized size. We compute size reduction as $100 \cdot \left(1 - \frac{S_q}{S_{\text{F16}}}\right)\%$.

\subsection{Evaluation Tasks \& Datasets}
\label{sec:setup:tasks}
To characterize the accuracy--efficiency trade-offs induced by \texttt{llama.cpp} quantization, we evaluate each model on a benchmark suite designed for broad coverage: multi-step math (GSM8K), commonsense multiple-choice reasoning (HellaSwag), instruction following (IFEval), broad knowledge (MMLU), truthfulness (TruthfulQA), and intrinsic language modeling quality (WikiText-2 Perplexity). Because strong base models typically perform well on these axes in full precision, this suite makes it easy to detect where quantization degrades capabilities (or leaves them largely intact), rather than relying on a single proxy task. 

We evaluate mathematical reasoning on GSM8K (task version v3) using 5-shot prompting. Commonsense reasoning is evaluated on HellaSwag (task version v1); instruction following is measured with IFEval (task version v4); broad knowledge is evaluated on MMLU (task version v2); and truthfulness is assessed with TruthfulQA. Finally, we report perplexity on WikiText-2 as a prompt-independent measure of language modeling quality. WikiText-2 is obtained via \texttt{llama.cpp} \texttt{./scripts/get-wikitext-2.sh}, and perplexity is computed with identical runtime settings across quantization schemes. We also record throughput under fixed settings of \(\texttt{pp}=512\) (prompt processing) and \(\texttt{tg}=128\) (token generation), enabling direct quality--speed comparisons.

The benchmark scores GSM8K, HellaSwag, IFEval, MMLU and TruthfulQA is evaluated with the LLM Evaluation Harness, while WikiText-2 perplexity and throughput are computed separately using \texttt{llama.cpp}'s evaluation script, since it provides the perplexity implementation used by \texttt{llama-bench} and \texttt{llama-perplexity} by \texttt{llama.cpp}. 

\section{Quantization Results and Trade-offs}
\label{sec:analysis}

\subsection{Overall Accuracy-Compression behavior}
\label{sec:analysis:overall}

Table~\ref{tab:full-results-wide} summarizes a clear but non-monotonic relationship between nominal bit-width, downstream accuracy, and intrinsic language-modeling quality. Complete evaluation suite with full metrics can be viewed in Appendix \ref{sec:appendixb}. As expected, the most aggressive 3-bit configurations deliver the largest space savings, but they also introduce the largest average degradation across the benchmark suite. The strongest compression point, \texttt{Q3\_K\_S}, achieves the highest reduction yet yields the largest drop in the unweighted benchmark mean (Avg), and it also exhibits the largest perplexity increase on WikiText-2. Moving within the 3-bit family (\texttt{Q3\_K\_M} and \texttt{Q3\_K\_L}) reduces the damage substantially, indicating that the implementation details of a quantization format (e.g., scaling granularity and block structure) matter at least as much as the headline ``3-bit'' label.

More interestingly, \texttt{Q5\_0} and \texttt{Q5\_1} quantization schemes slightly exceed the \texttt{F16} baseline on the downstream mean under our fixed decoding and evaluation protocol. In Table~\ref{tab:full-results-wide}, \texttt{Q5\_0} attains the highest Avg among all configurations while still reducing model size by roughly sixty-five percent. These small gains over \texttt{F16} should be interpreted cautiously. Even with deterministic decoding, finite benchmark sets introduce evaluation variance, and modest improvements can reflect noise or idiosyncrasies of the scoring pipeline rather than true general superiority. The robust takeaway is that a carefully chosen mid-bit quantization can preserve the baseline capability envelope surprisingly well while dramatically reducing storage and improving throughput.

\begin{table*}[t]
\centering
\caption{Full downstream and perplexity results for \texttt{llama3.1-8b} under \texttt{llama.cpp} quantization schemes.}
\label{tab:full-results-wide}

\begin{tabular}{llcccccccc}
\toprule
& & & \multicolumn{6}{c}{Benchmarks ($\uparrow$)} & \\
\cmidrule(lr){4-9}
Bits & Quant & Size Reduction (\%) & GSM8K & HSwag & IFEval & MMLU & TQA$_{\text{mc2}}$ & Avg & PPL $\downarrow$ \\
\midrule
\multicolumn{2}{l}{F16 (baseline)} & -- &
77.63 & 72.51 & 78.93 & 63.50 & 54.79 & 69.47 & 7.32 \\
\midrule
3 & Q3\_K\_S & 77.23 & 68.31 & 71.87 & 73.89 & 59.31 & 54.08 & 65.49 & 8.96 \\
  & Q3\_K\_M & 75.03 & 73.16 & 73.41 & 77.19 & 62.01 & 54.56 & 68.07 & 7.96 \\
  & Q3\_K\_L & 73.14 & 74.07 & 73.54 & 79.14 & 62.31 & 54.84 & 68.78 & 7.81 \\
\midrule
4 & Q4\_0    & 71.03 & 75.66 & 71.88 & 77.46 & 62.20 & 52.68 & 67.98 & 7.74 \\
  & Q4\_1    & 68.11 & 76.04 & 71.29 & 78.45 & 63.17 & 55.01 & 68.79 & 7.72 \\
  & Q4\_K\_S & 70.83 & 77.33 & 72.79 & 80.26 & 62.06 & 53.40 & 69.17 & 7.62 \\
  & Q4\_K\_M & 69.41 & 77.41 & 72.35 & 79.06 & 62.43 & 54.49 & 69.15 & 7.56 \\
\midrule
5 & Q5\_0    & 65.19 & 79.08 & 72.63 & 80.14 & 63.18 & 54.57 & 69.92 & 7.43 \\
  & Q5\_1    & 62.27 & 78.47 & 72.08 & 79.79 & 63.68 & 54.62 & 69.73 & 7.43 \\
  & Q5\_K\_S & 65.19 & 75.66 & 72.67 & 79.50 & 63.36 & 53.90 & 69.02 & 7.43 \\
  & Q5\_K\_M & 64.35 & 78.54 & 72.33 & 78.67 & 62.80 & 54.45 & 69.36 & 7.40 \\
\midrule
6 & Q6\_K    & 58.98 & 78.17 & 72.48 & 77.63 & 63.17 & 54.71 & 69.23 & 7.35 \\
\midrule
8 & Q8\_0    & 46.87 & 77.48 & 72.52 & 78.79 & 63.43 & 54.81 & 69.41 & 7.33 \\
\bottomrule
\end{tabular}

\vspace{0.5ex}
\begin{minipage}{0.98\linewidth}
\footnotesize
GSM8K: flexible-extract. HSwag: acc\_norm). IFEval: mean of four accuracies (instruction/prompt $\times$ loose/strict). MMLU: acc. TQA$_{\text{mc2}}$: MC2 accuracy. Avg: unweighted mean of the five benchmark scores (GSM8K, HSwag, IFEval, MMLU, TQA$_{\text{mc2}}$), excluding PPL. PPL: WikiText-2 perplexity (lower is better). Cells show means rounded to 2 decimals; standard errors and additional breakdowns are reported in Appendix \ref{sec:appendixb}.
\end{minipage}
\end{table*}

\subsection{Task Sensitivity Under Quantization}
\label{sec:analysis:task}
The per-benchmark columns in Table~\ref{tab:full-results-wide} make clear that quantization reshapes model behavior in a task-dependent way rather than inducing a uniform accuracy shift. Multi-step arithmetic reasoning on GSM8K is among the most sensitive dimensions: relative to the \texttt{F16} baseline of $77.63$, the most aggressive 3-bit configuration \texttt{Q3\_K\_S} drops to $68.31$ (a $-9.32$ point change), while moving within the 3-bit family already recovers a substantial fraction of the loss, reaching $73.16$ for \texttt{Q3\_K\_M} and $74.07$ for \texttt{Q3\_K\_L}. At 4 bits, the gap largely closes, with \texttt{Q4\_0} and \texttt{Q4\_1} scoring $75.66$ and $76.04$, and the \texttt{K}-quantized variants essentially matching baseline at $77.33$ (\texttt{Q4\_K\_S}) and $77.41$ (\texttt{Q4\_K\_M}). Several 5-bit and higher settings even exceed \texttt{F16} on GSM8K, including \texttt{Q5\_0} at $79.08$, \texttt{Q5\_1} at $78.47$, \texttt{Q5\_K\_M} at $78.54$, and \texttt{Q6\_K} at $78.17$, while \texttt{Q5\_K\_S} is a notable exception that falls back to $75.66$ despite sharing the same nominal bit-width. This non-monotonicity indicates that GSM8K is sensitive not just to bit-width, but to the particular quantization format and its induced error structure.

In contrast, commonsense multiple-choice accuracy on HellaSwag is comparatively stable across the entire sweep. The \texttt{F16} baseline is $72.51$, and most quantized variants remain within roughly a one-point neighborhood: the lowest value appears at \texttt{Q4\_1} with $71.29$, whereas the strongest scores slightly exceed baseline, such as \texttt{Q3\_K\_L} at $73.54$ and \texttt{Q3\_K\_M} at $73.41$. Even the 8-bit entry \texttt{Q8\_0} essentially reproduces baseline at $72.52$. This small variation, even with large changes in compression ratio, suggests that HellaSwag performance is largely unaffected by the noise introduced by these quantization schemes.

Instruction-following robustness, as measured by IFEval, sits between these extremes. While \texttt{Q3\_K\_S} shows a sizable regression from $78.93$ (\texttt{F16}) down to $73.89$, several mid-bit settings are not merely resilient but improve over baseline. Notably, \texttt{Q3\_K\_L} reaches $79.14$, \texttt{Q4\_K\_S} increases to $80.26$, and \texttt{Q5\_0} attains $80.14$, indicating that moderate quantization can preserve (and occasionally enhance) instruction-following outcomes. At the same time, improvements are not guaranteed by higher precision: \texttt{Q6\_K} drops to $77.63$ and \texttt{Q8\_0} remains close to baseline at $78.79$, reinforcing that IFEval is sensitive to format details rather than bit-width alone.

Broad knowledge on MMLU shows moderate sensitivity with a clear failure mode at the most aggressive setting but otherwise small fluctuations. From a baseline of $63.50$, \texttt{Q3\_K\_S} declines to $59.31$, yet most other configurations stay within about a point of \texttt{F16}. For example, \texttt{Q4\_1} and \texttt{Q6\_K} both score $63.17$, \texttt{Q8\_0} reaches $63.43$, and \texttt{Q5\_1} slightly surpasses baseline at $63.68$. 

For TruthfulQA, although \texttt{Q4\_0} dips to $52.68$ versus $54.79$ for \texttt{F16}, many formats remain close, including \texttt{Q4\_1} at $55.01$ and \texttt{Q8\_0} at $54.81$. 

These benchmark-specific behaviors are reflected in the overall average as well: \texttt{Q3\_K\_S} produces the largest aggregate drop (Avg $65.49$ vs.\ $69.47$), while most 4--8 bit schemes cluster tightly near baseline, and \texttt{Q5\_0} achieves the highest mean at $69.92$.

Finally, the perplexity column provides a useful complementary signal about how aggressively each scheme perturbs the model distribution. \texttt{F16} yields $7.32$, whereas the most aggressive 3-bit setting \texttt{Q3\_K\_S} increases perplexity to $8.96$, broadly aligning with its across-task accuracy losses. However, perplexity is not a complete predictor of downstream behavior: several 5-bit variants share essentially the same PPL (e.g., $7.43$ for \texttt{Q5\_0}, \texttt{Q5\_1}, and \texttt{Q5\_K\_S}) while exhibiting markedly different GSM8K and IFEval outcomes, underscoring that task sensitivity depends on how quantization error interacts with the computation pattern demanded by a benchmark rather than on distributional fit alone.

\subsection{Pareto Frontier Analysis: Size Reduction and Performance Tradeoff}
\label{sec:analysis:pareto}
\begin{figure}[htbp]
  \centering
  \includegraphics[width=\linewidth]{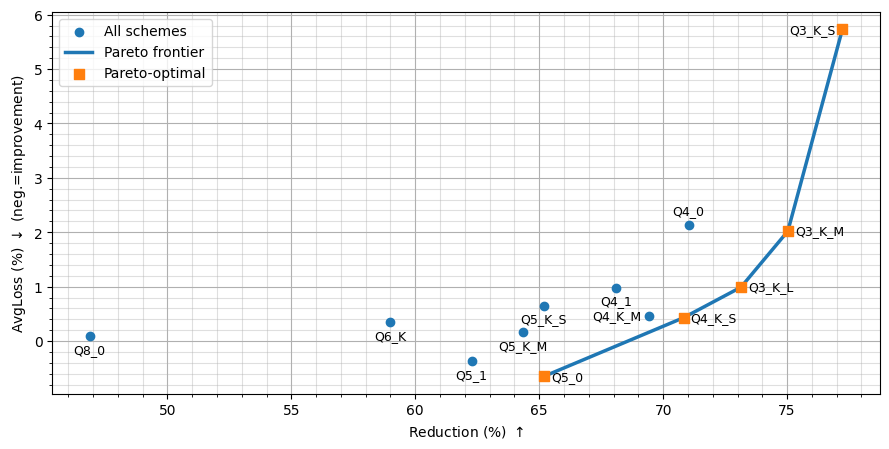}
  \caption{Compression vs benchmark quality loss. AvgLoss(\%) $=\frac{\mathrm{Avg}_{\text{F16}}-\mathrm{Avg}}{\mathrm{Avg}_{\text{F16}}}\times 100$.}
  \label{pareto}
\end{figure}

Figure \ref{pareto} summarizes the compression--quality trade-off using two derived metrics from Table \ref{tab:full-results-wide}. The x-axis reports reduction as the compression ratio (\%), and the y-axis reports AvgLoss (\%) computed from the table \ref{tab:full-results-wide}'s Avg column as a relative change with respect to the \texttt{F16} baseline, i.e., $\text{AvgLoss} = 100 \cdot (\text{Avg}_{\text{F16}} - \text{Avg})/\text{Avg}_{\text{F16}}$, so that negative values correspond to a slight improvement over baseline. Under this two-objective view, a configuration is Pareto-optimal when no other configuration simultaneously achieves higher reduction and lower AvgLoss, and the Pareto frontier therefore enumerates the operating points that remain defensible once both memory savings and aggregate downstream quality are considered.

Reading the frontier from the quality-preserving end, \texttt{Q5\_0} is the most favorable point in terms of AvgLoss, achieving a $65.19\%$ reduction while slightly improving the aggregate mean (AvgLoss $\approx -0.65\%$). This is not merely a marginal win. It also removes several nearby options from consideration because they are strictly dominated under the two metrics, including \texttt{Q5\_1}, \texttt{Q5\_K\_S}, and \texttt{Q5\_K\_M}, all of which deliver less reduction and higher AvgLoss than \texttt{Q5\_0}. When additional compression is required, the frontier moves to \texttt{Q4\_K\_S}, which increases reduction to $70.83\%$ while keeping the quality penalty small (AvgLoss $\approx 0.43\%$). This point captures a strong trade-off within the 4-bit family, and it dominates plausible alternatives such as \texttt{Q4\_K\_M} and \texttt{Q4\_1} by offering both higher reduction and lower AvgLoss. 

Pushing further toward maximum compression, \texttt{Q3\_K\_L} reaches $73.14\%$ reduction with AvgLoss $\approx 0.99\%$, representing a modest additional compression gain for a still-controlled aggregate degradation. \texttt{Q3\_K\_M} attains a $75.03\%$ reduction with AvgLoss $\approx 2.02\%$, and becomes Pareto-optimal because no other evaluated scheme achieves at least as much reduction with lower AvgLoss. At the extreme end, \texttt{Q3\_K\_S} attains the largest reduction in the table ($77.23\%$) and remains Pareto-optimal only because no other scheme matches its reduction, but this comes with a substantially larger quality cost (AvgLoss $\approx 5.73\%$), making it appropriate primarily when memory footprint is the overriding constraint.

This Pareto framing also makes clear why more bits are not automatically better once compression is explicitly valued. Schemes like \texttt{Q6\_K} and \texttt{Q8\_0} stay close to baseline quality, but they are nevertheless dominated because \texttt{Q5\_0} provides both higher reduction and lower AvgLoss. In practical terms, the efficient set implied by Figure~\ref{pareto} is compact: \texttt{Q5\_0} is the accuracy-favoring choice with meaningful compression, \texttt{Q4\_K\_S} is the natural balanced default when stronger reduction is needed, \texttt{Q3\_K\_L} offers a further compression step with moderate additional loss, \texttt{Q3\_K\_M} provides an additional compression step at a higher but still controlled aggregate loss, and \texttt{Q3\_K\_S} is the maximum-reduction endpoint for the most memory-constrained deployments.

\subsection{Throughput Analysis}
\label{sec:analysis:throughput}
Table~\ref{tab:systems_throughput} complements the accuracy results with throughput measured under standardized settings (\texttt{pp}=512, \texttt{tg}=128). Token-generation speed (tg128) increases substantially for lower-bit schemes, with the most compressed formats producing the largest gains. This behavior is consistent with quantization reducing memory bandwidth pressure and improving cache residency, which are often limiting factors in CPU inference. The most compressed configuration achieves the highest token-generation throughput, making it attractive for interactive settings where latency per generated token dominates.

Prompt-processing throughput (pp512) varies more irregularly. Unlike pure decoding, prompt processing can be affected by different kernel choices, memory access patterns, and the balance between compute and bandwidth across formats. The key operational point is that throughput improvements are not solely determined by nominal bit-width, and measuring both pp and tg under a fixed configuration is important when the application alternates between long prompts and long generations.

When we combine throughput with the Pareto analysis, \texttt{Q4\_K\_S} again appears as a strong default. It sits on the non-dominated accuracy--compression frontier while delivering a large generation-speed gain relative to \texttt{F16}. In contrast, \texttt{Q3\_K\_S} maximizes throughput and compression, but the associated accuracy loss is large enough that it is best treated as a specialized choice for constrained environments rather than a general-purpose setting.

\begin{table}[t]
\centering
\begin{tabular}{lcc}
\toprule
Scheme & pp512 $\uparrow$ & tg128 $\uparrow$ \\
\midrule
F16       & 79.57 $\pm$ 1.00 & 2.83 $\pm$ 0.01 \\
Q3\_K\_S  & 57.39 $\pm$ 3.01 & 9.91 $\pm$ 0.03 \\
Q3\_K\_M  & 68.86 $\pm$ 1.36 & 6.52 $\pm$ 0.23 \\
Q3\_K\_L  & 58.34 $\pm$ 1.93 & 7.93 $\pm$ 0.06 \\
Q4\_0     & 97.35 $\pm$ 0.68 & 4.36 $\pm$ 0.20 \\
Q4\_1     & 55.20 $\pm$ 1.61 & 6.80 $\pm$ 0.15 \\
Q4\_K\_S  & 92.52 $\pm$ 1.53 & 4.65 $\pm$ 0.15 \\
Q4\_K\_M  & 87.70 $\pm$ 0.70 & 5.12 $\pm$ 0.37 \\
Q5\_0     & 61.44 $\pm$ 2.47 & 6.66 $\pm$ 0.06 \\
Q5\_1     & 45.98 $\pm$ 2.17 & 6.33 $\pm$ 0.26 \\
Q5\_K\_S  & 55.31 $\pm$ 1.67 & 5.98 $\pm$ 0.02 \\
Q5\_K\_M  & 58.24 $\pm$ 2.05 & 6.85 $\pm$ 0.13 \\
Q6\_K     & 59.81 $\pm$ 1.07 & 6.33 $\pm$ 0.13 \\
Q8\_0     & 71.42 $\pm$ 3.15 & 5.03 $\pm$ 0.16 \\
\bottomrule
\end{tabular}
\caption{Throughput metrics (pp512, tg128) by scheme. Throughputs in tokens/sec for prompt processing of 512 tokens and generation of 128 tokens (wall-clock), and the configuration shown in Section \ref{sec:setup:base}}
\label{tab:systems_throughput}
\end{table}

\subsection{Choosing Quantization Scheme In Practice}
Table~4 is intended as a quick, deployment-oriented guide for selecting a quantization format without having to scan the full set of results. When memory and disk are the primary bottlenecks, lower-bit K-quants deliver the largest reductions and strong speedups, but Table~4 makes clear that the most aggressive settings also incur the largest perplexity increases and average benchmark drops.

For general-purpose interactive use, the table points to mid-bit formats as sensible defaults: 4-bit K-quants typically offer near-maximal compression at 4-bit while keeping task performance close to the FP16 baseline, and a high-quality 5-bit option is a common step up when additional quality margin is needed. When preserving behavior is the priority (e.g., instruction following or reasoning sensitivity), Table~4 favors these mid-bit regimes over very low-bit choices, while conservative 6--8 bit settings remain appropriate when perplexity and downstream drift must be minimized, and memory budgets allow.


\setlength{\LTleft}{0pt}
\setlength{\LTright}{0pt}

\begin{longtable}{@{}
P{\dimexpr0.18\linewidth-2\tabcolsep\relax}
P{\dimexpr0.21\linewidth-2\tabcolsep\relax}
P{\dimexpr0.25\linewidth-2\tabcolsep\relax}
P{\dimexpr0.36\linewidth-2\tabcolsep\relax}
@{}}
\label{tab:scenario-regimes}\\

\toprule
Scenario & Primary constraint / goal & Deployable regimes (examples) & Trade-offs (from your results) \\
\midrule
\endfirsthead

\toprule
Scenario & Primary constraint / goal & Deployable regimes (examples) & Trade-offs (from your results) \\
\midrule
\endhead

\midrule
\multicolumn{4}{r}{\emph{Continued on next page}}\\
\endfoot

\bottomrule
\endlastfoot

Edge / disk- or RAM-limited serving &
Minimize footprint; accept visible quality loss &
\textbf{3-bit (ultra-compressed)}: \qcode{Q3_K_S}, \qcode{Q3_K_L} &
Biggest size reduction and very strong tg speedups, but the largest Avg drops and highest PPL increases (especially \qcode{Q3_K_S}); \qcode{Q3_K_M} is the safer 3-bit choice. \\
\midrule

Interactive chat on CPU (general-purpose) &
Best overall balance of size, quality, and speed &
\textbf{4--5 bit (balanced default)}: \qcode{Q4_K_S};
also \qcode{Q4_0}, \qcode{Q4_K_M};
or \qcode{Q5_0} if size allows &
\qcode{Q4_K_S} gives near-max 4-bit compression with Avg close to F16 and solid tg gains; \qcode{Q5_0} is larger but delivers the strongest Avg and faster tg. \\
\midrule

Throughput-first serving (quality-aware) &
Max tokens/s at acceptable quality &
\textbf{3--5 bit}: \qcode{Q3_K_M} (if loss is OK),
\qcode{Q5_0}/\qcode{Q5_K_M} (safer),
\qcode{Q4_K_S} (more compression) &
3-bit maximizes speed but can hurt math/instruction tasks; \qcode{Q5_0}/\qcode{Q5_K_M} keep quality near (or above) baseline while staying very fast; \qcode{Q4_K_S} trades some throughput for more compression. \\
\midrule

Accuracy-first CPU deployment (not full F16) &
Minimize behavior drift vs baseline; accept larger model &
\textbf{6--8 bit (conservative)}: \qcode{Q6_K}, \qcode{Q8_0} &
Perplexity is closest to F16 and downstream deltas are small, but compression is weaker than the best 4--5 bit options. \\
\midrule

Math-/reasoning-heavy workloads &
Protect GSM8K-style multi-step performance &
\textbf{5-bit or strong 4-bit}: \qcode{Q5_0} (also \qcode{Q5_1}), \qcode{Q4_K_S} &
Avoid aggressive 3-bit (notably \qcode{Q3_K_S}) which shows the largest GSM8K drop; \qcode{Q5_0} and \qcode{Q4_K_S} preserve downstream performance with substantial compression. \\
\midrule

Instruction-following sensitive applications &
Protect IFEval-like compliance &
\textbf{4--5 bit}: \qcode{Q4_K_S}, \qcode{Q5_0} (also \qcode{Q4_K_M}) &
IFEval stays strong for several mid-bit schemes; 3-bit is more likely to reduce compliance. \\
\midrule

Calibration / ``LM quality'' sensitive (proxy via PPL) &
Keep perplexity close to baseline &
\textbf{6--8 bit} (or \textbf{good 5-bit}):
\qcode{Q6_K}, \qcode{Q8_0}, \qcode{Q5_K_M} (or \qcode{Q5_0}) &
PPL generally worsens as compression increases; if PPL is a hard guardrail, prefer Q6/Q8, with a 5-bit option (e.g., \qcode{Q5_K_M} / \qcode{Q5_0}) as a middle ground. \\

\caption{Deployment scenarios and which quantization regimes are practically deployable under the measured trade-offs. Deployable here means the scheme is a sensible candidate given the primary constraint; final choice should still be validated on the target workload.}
\end{longtable}

\section{Conclusion}
This paper provides a unified empirical comparison of widely used \texttt{llama.cpp} GGUF quantization formats for a modern instruction-tuned model under a single controlled pipeline, reporting downstream task performance alongside intrinsic quality, footprint, and practical efficiency. The results show that mid-bit quantization can preserve capabilities surprisingly well. In our setting, 5-bit configurations offer an especially strong trade-off, maintaining (and in some cases slightly improving) aggregate downstream scores relative to the FP16 baseline while substantially reducing model size, whereas the most aggressive 3-bit setting exhibits the clearest and most consistent degradations, particularly on reasoning-heavy evaluations.

A key takeaway is that the quantization format matters, not just the nominal bit-width: schemes with similar perplexity can diverge meaningfully on instruction-following and reasoning benchmarks, so intrinsic metrics alone are insufficient to select deployment defaults. These findings motivate a simple evidence-based guidance: prefer high-performing 5-bit options as a robust default when memory is limited but quality is important, consider well-tuned 4-bit K-quant variants when footprint must be pushed further with modest loss, and treat 3-bit choices as budget-driven compromises. Future work should extend the same protocol across additional model families and sizes, diverse hardware targets, and longer-context regimes, and should analyze how format design choices mechanistically relate to the observed task-specific regressions.

\FloatBarrier
\bibliographystyle{unsrt}  
\bibliography{references}  

\appendix
\section{Quantization Timings}
\label{sec:appendix-a}

Table \ref{tab:quant-configs} summarizes the deployment-side properties of each GGUF quantization scheme evaluated in this study: the resulting model size (MiB), the corresponding size reduction relative to the FP16 GGUF input, and the wall-clock time required to perform the quantization with the official llama.cpp tooling on our evaluation system. These measurements complement the main-text quality and throughput results by making explicit the storage savings and one-time conversion overhead associated with each format, which are often decisive constraints in CPU and edge-oriented deployments.
\begin{table}[t]
\centering
\begin{tabular}{lrrrr}
\toprule
Scheme & Bits & Size (MiB) & Size reduction (\%) & Quant time (s) \\
\midrule
Q3\_K\_S & 3 & 3487.27 & 77.23 & 27.04 \\
Q3\_K\_M & 3 & 3825.27 & 75.03 & 31.82 \\
Q3\_K\_L & 3 & 4114.27 & 73.14 & 31.37 \\
Q4\_0    & 4 & 4437.80 & 71.03 & 27.35 \\
Q4\_1    & 4 & 4885.12 & 68.11 & 27.76 \\
Q4\_K\_S & 4 & 4467.80 & 70.83 & 42.84 \\
Q4\_K\_M & 4 & 4685.30 & 69.41 & 42.19 \\
Q5\_0    & 5 & 5332.43 & 65.19 & 28.21 \\
Q5\_1    & 5 & 5779.74 & 62.27 & 29.46 \\
Q5\_K\_S & 5 & 5332.43 & 65.19 & 37.47 \\
Q5\_K\_M & 5 & 5459.93 & 64.35 & 36.47 \\
Q6\_K    & 6 & 6282.97 & 58.98 & 31.19 \\
Q8\_0    & 8 & 8137.64 & 46.87 & 29.26 \\
\bottomrule
\end{tabular}
\caption{Quantization configurations and costs. Size reductions are relative to the $S_{\text{F16}}=15317.02\,\mathrm{MiB}$ GGUF input.}
\label{tab:quant-configs}
\end{table}

\FloatBarrier
\section{Full Quantization Results}
\label{sec:appendixb}
Table \ref{tab:benchmark-quant} reports the complete set of benchmark outputs underlying the aggregated scores in the main results table, including mean performance (and where applicable, standard error) for each quantization type on every reported task/metric. In addition to overall scores, the table includes alternative scoring variants (e.g., two GSM8K exact-match criteria), multiple views of instruction-following for IFEval (instruction-level vs prompt-level, each under loose vs strict evaluation), MMLU broken down into subject areas, and TruthfulQA reported for multiple-choice variant.

The accuracy/metric abbreviations denote the specific scoring variants used by the evaluation harness. For GSM8K, the reported metric is EM (Exact Match), i.e., the percentage of problems where the model’s final answer exactly matches the gold answer after extraction; FE (Flexible Extract) and SM (Strict Match) specify the extraction filter used before computing EM (FE is lenient in extracting the final numeric answer from text, while SM is stricter about the expected final-answer format). For HellaSwag, A indicates raw multiple-choice acc (accuracy) and AN indicates acc\_norm, i.e., normalized accuracy. For IFEval, ILL/ILS correspond to instruction-level loose/strict accuracy (inst\_level\_loose\_acc / inst\_level\_strict\_acc) and PLL/PLS to prompt-level loose/strict accuracy (prompt\_level\_loose\_acc / prompt\_level\_strict\_acc). For MMLU, A denotes overall acc (accuracy), with additional rows reporting subject-group accuracies. And for TruthfulQA, the generation setting reports BA/BD as bleu\_acc (BLEU-based accuracy) and bleu\_dif (BLEU-based difference), while the multiple-choice settings MC1 and MC2 report A as acc (accuracy).

Unless otherwise noted, the benchmarks were run under the following harness versions and shot settings: GSM8K v3 (5-shot), HellaSwag v1 (0-shot), IFEval v4 (0-shot), MMLU v2 (0-shot), and TruthfulQA v3 (0-shot), with the exception that TruthfulQA MC1 uses v2.

\setlength{\LTcapwidth}{\textwidth}
\setlength{\LTleft}{\fill}
\setlength{\LTright}{\fill}

\begin{longtable}{llcr}
\toprule
QType & Task & Metric & Value\\
\midrule




F16 (baseline) & GSM8K (FE) & EM  & 77.63 $\pm$ 1.15\\
               & GSM8K (SM) & EM  & 24.64 $\pm$ 1.19\\
               & HellaSwag & A   & 57.40 $\pm$ 0.49\\
               & HellaSwag & AN  & 72.51 $\pm$ 0.45\\
               & IFEval & ILL & 83.81\\
               & IFEval & ILS & 81.06\\
               & IFEval & PLL & 77.08 $\pm$ 1.81\\
               & IFEval & PLS & 73.75 $\pm$ 1.89\\
               & MMLU & A   & 63.50 $\pm$ 0.38\\
               & MMLU Humanities & A   & 59.21 $\pm$ 0.68\\
               & MMLU Other & A   & 71.97 $\pm$ 0.78\\
               & MMLU Social Sciences & A   & 74.59 $\pm$ 0.77\\
               & MMLU Stem & A   & 50.71 $\pm$ 0.84\\
               & TruthfulQA Gen & BA  & 46.27 $\pm$ 1.75\\
               & TruthfulQA Gen & BD  & -22.91 $\pm$ 23.06\\
               & TruthfulQA MC1 & A   & 39.53 $\pm$ 1.71\\
               & TruthfulQA MC2 & A   & 54.79 $\pm$ 1.60\\
\midrule
\midrule

Q3\_K\_S & GSM8K (FE) & EM  & 68.31 $\pm$ 1.28\\
      & GSM8K (SM) & EM  & 22.14 $\pm$ 1.14\\
      & HellaSwag & A   & 56.54 $\pm$ 0.49\\
      & HellaSwag & AN  & 71.87 $\pm$ 0.45\\
      & IFEval & ILL & 79.50\\
      & IFEval & ILS & 76.86\\
      & IFEval & PLL & 71.16 $\pm$ 1.95\\
      & IFEval & PLS & 68.02 $\pm$ 2.01\\
      & MMLU & A   & 59.31 $\pm$ 0.39\\
      & MMLU Humanities & A   & 52.67 $\pm$ 0.68\\
      & MMLU Other & A   & 67.81 $\pm$ 0.82\\
      & MMLU Social Sciences & A   & 70.07 $\pm$ 0.81\\
      & MMLU Stem & A   & 50.36 $\pm$ 0.86\\
      & TruthfulQA Gen & BA  & 47.37 $\pm$ 1.75\\
      & TruthfulQA Gen & BD  & -30.76 $\pm$ 32.44\\
      & TruthfulQA MC1 & A   & 36.96 $\pm$ 1.69\\
      & TruthfulQA MC2 & A   & 54.08 $\pm$ 1.58\\
\midrule

Q3\_K\_M & GSM8K (FE) & EM  & 73.16 $\pm$ 1.22\\
      & GSM8K (SM) & EM  & 9.86 $\pm$ 0.82\\
      & HellaSwag & A   & 57.62 $\pm$ 0.49\\
      & HellaSwag & AN  & 73.41 $\pm$ 0.44\\
      & IFEval & ILL & 82.49\\
      & IFEval & ILS & 79.14\\
      & IFEval & PLL & 75.97 $\pm$ 1.84\\
      & IFEval & PLS & 71.16 $\pm$ 1.95\\
      & MMLU & A   & 62.01 $\pm$ 0.39\\
      & MMLU Humanities & A   & 58.04 $\pm$ 0.69\\
      & MMLU Other & A   & 68.81 $\pm$ 0.81\\
      & MMLU Social Sciences & A   & 71.60 $\pm$ 0.79\\
      & MMLU Stem & A   & 51.86 $\pm$ 0.85\\
      & TruthfulQA Gen & BA  & 43.57 $\pm$ 1.74\\
      & TruthfulQA Gen & BD  & -40.98 $\pm$ 33.19\\
      & TruthfulQA MC1 & A   & 39.53 $\pm$ 1.71\\
      & TruthfulQA MC2 & A   & 54.56 $\pm$ 1.58\\
\midrule

Q3\_K\_L & GSM8K (FE) & EM  & 74.07 $\pm$ 1.21\\
      & GSM8K (SM) & EM  & 10.54 $\pm$ 0.85\\
      & HellaSwag & A   & 57.28 $\pm$ 0.49\\
      & HellaSwag & AN  & 73.54 $\pm$ 0.44\\
      & IFEval & ILL & 84.41\\
      & IFEval & ILS & 80.94\\
      & IFEval & PLL & 77.82 $\pm$ 1.79\\
      & IFEval & PLS & 73.38 $\pm$ 1.90\\
      & MMLU & A   & 62.31 $\pm$ 0.39\\
      & MMLU Humanities & A   & 58.75 $\pm$ 0.69\\
      & MMLU Other & A   & 69.68 $\pm$ 0.80\\
      & MMLU Social Sciences & A   & 71.08 $\pm$ 0.80\\
      & MMLU Stem & A   & 51.79 $\pm$ 0.85\\
      & TruthfulQA Gen & BA  & 41.74 $\pm$ 1.73\\
      & TruthfulQA Gen & BD  & -36.16 $\pm$ 27.06\\
      & TruthfulQA MC1 & A   & 39.05 $\pm$ 1.71\\
      & TruthfulQA MC2 & A   & 54.84 $\pm$ 1.58\\
\midrule
\midrule

Q4\_0 & GSM8K (FE) & EM  & 75.66 $\pm$ 1.18\\
     & GSM8K (SM) & EM  & 18.95 $\pm$ 1.08\\
     & HellaSwag & A   & 56.76 $\pm$ 0.49\\
     & HellaSwag & AN  & 71.88 $\pm$ 0.45\\
     & IFEval & ILL & 83.09\\
     & IFEval & ILS & 79.98\\
     & IFEval & PLL & 75.42 $\pm$ 1.85\\
     & IFEval & PLS & 71.35 $\pm$ 1.95\\
     & MMLU & A   & 62.20 $\pm$ 0.39\\
     & MMLU Humanities & A   & 57.56 $\pm$ 0.69\\
     & MMLU Other & A   & 69.91 $\pm$ 0.80\\
     & MMLU Social Sciences & A   & 72.93 $\pm$ 0.78\\
     & MMLU Stem & A   & 51.06 $\pm$ 0.85\\
     & TruthfulQA Gen & BA  & 52.88 $\pm$ 1.75\\
     & TruthfulQA Gen & BD  & 30.93 $\pm$ 26.02\\
     & TruthfulQA MC1 & A   & 38.31 $\pm$ 1.70\\
     & TruthfulQA MC2 & A   & 52.68 $\pm$ 1.60\\
\midrule

Q4\_1 & GSM8K (FE) & EM  & 76.04 $\pm$ 1.18\\
     & GSM8K (SM) & EM  & 23.58 $\pm$ 1.17\\
     & HellaSwag & A   & 56.28 $\pm$ 0.50\\
     & HellaSwag & AN  & 71.29 $\pm$ 0.45\\
     & IFEval & ILL & 84.05\\
     & IFEval & ILS & 80.22\\
     & IFEval & PLL & 77.26 $\pm$ 1.80\\
     & IFEval & PLS & 72.27 $\pm$ 1.93\\
     & MMLU & A   & 63.17 $\pm$ 0.39\\
     & MMLU Humanities & A   & 58.94 $\pm$ 0.69\\
     & MMLU Other & A   & 70.68 $\pm$ 0.79\\
     & MMLU Social Sciences & A   & 72.93 $\pm$ 0.78\\
     & MMLU Stem & A   & 52.55 $\pm$ 0.85\\
     & TruthfulQA Gen & BA  & 44.80 $\pm$ 1.74\\
     & TruthfulQA Gen & BD  & -33.30 $\pm$ 31.46\\
     & TruthfulQA MC1 & A   & 38.56 $\pm$ 1.70\\
     & TruthfulQA MC2 & A   & 55.01 $\pm$ 1.61\\
\midrule

Q4\_K\_S & GSM8K (FE) & EM  & 77.33 $\pm$ 1.15\\
      & GSM8K (SM) & EM  & 17.29 $\pm$ 1.04\\
      & HellaSwag & A   & 57.56 $\pm$ 0.49\\
      & HellaSwag & AN  & 72.79 $\pm$ 0.44\\
      & IFEval & ILL & 85.01\\
      & IFEval & ILS & 81.89\\
      & IFEval & PLL & 79.11 $\pm$ 1.75\\
      & IFEval & PLS & 75.05 $\pm$ 1.86\\
      & MMLU & A   & 62.06 $\pm$ 0.39\\
      & MMLU Humanities & A   & 58.17 $\pm$ 0.69\\
      & MMLU Other & A   & 69.58 $\pm$ 0.80\\
      & MMLU Social Sciences & A   & 71.60 $\pm$ 0.79\\
      & MMLU Stem & A   & 51.13 $\pm$ 0.85\\
      & TruthfulQA Gen & BA  & 51.77 $\pm$ 1.75\\
      & TruthfulQA Gen & BD  & 12.04 $\pm$ 25.97\\
      & TruthfulQA MC1 & A   & 37.45 $\pm$ 1.69\\
      & TruthfulQA MC2 & A   & 53.40 $\pm$ 1.59\\
\midrule

Q4\_K\_M & GSM8K (FE) & EM  & 77.41 $\pm$ 1.15\\
      & GSM8K (SM) & EM  & 14.48 $\pm$ 0.97\\
      & HellaSwag & A   & 57.26 $\pm$ 0.49\\
      & HellaSwag & AN  & 72.35 $\pm$ 0.45\\
      & IFEval & ILL & 84.05\\
      & IFEval & ILS & 80.82\\
      & IFEval & PLL & 77.63 $\pm$ 1.79\\
      & IFEval & PLS & 73.75 $\pm$ 1.89\\
      & MMLU & A   & 62.43 $\pm$ 0.39\\
      & MMLU Humanities & A   & 58.98 $\pm$ 0.69\\
      & MMLU Other & A   & 69.49 $\pm$ 0.80\\
      & MMLU Social Sciences & A   & 72.41 $\pm$ 0.79\\
      & MMLU Stem & A   & 50.87 $\pm$ 0.85\\
      & TruthfulQA Gen & BA  & 47.12 $\pm$ 1.75\\
      & TruthfulQA Gen & BD  & -17.69 $\pm$ 27.79\\
      & TruthfulQA MC1 & A   & 37.45 $\pm$ 1.69\\
      & TruthfulQA MC2 & A   & 54.49 $\pm$ 1.60\\
\midrule
\midrule

Q5\_0 & GSM8K (FE) & EM  & 79.08 $\pm$ 1.12\\
     & GSM8K (SM) & EM  & 37.68 $\pm$ 1.33\\
     & HellaSwag & A   & 57.41 $\pm$ 0.49\\
     & HellaSwag & AN  & 72.63 $\pm$ 0.44\\
     & IFEval & ILL & 85.01\\
     & IFEval & ILS & 82.13\\
     & IFEval & PLL & 78.56 $\pm$ 1.77\\
     & IFEval & PLS & 74.86 $\pm$ 1.87\\
     & MMLU & A   & 63.18 $\pm$ 0.38\\
     & MMLU Humanities & A   & 58.87 $\pm$ 0.69\\
     & MMLU Other & A   & 71.55 $\pm$ 0.78\\
     & MMLU Social Sciences & A   & 73.87 $\pm$ 0.77\\
     & MMLU Stem & A   & 50.94 $\pm$ 0.84\\
     & TruthfulQA Gen & BA  & 49.08 $\pm$ 1.75\\
     & TruthfulQA Gen & BD  & 3.46 $\pm$ 32.99\\
     & TruthfulQA MC1 & A   & 39.17 $\pm$ 1.71\\
     & TruthfulQA MC2 & A   & 54.57 $\pm$ 1.60\\
\midrule

Q5\_1 & GSM8K (FE) & EM  & 78.47 $\pm$ 1.13\\
     & GSM8K (SM) & EM  & 20.62 $\pm$ 1.11\\
     & HellaSwag & A   & 56.91 $\pm$ 0.49\\
     & HellaSwag & AN  & 72.08 $\pm$ 0.45\\
     & IFEval & ILL & 84.53\\
     & IFEval & ILS & 81.77\\
     & IFEval & PLL & 78.19 $\pm$ 1.78\\
     & IFEval & PLS & 74.68 $\pm$ 1.87\\
     & MMLU & A   & 63.68 $\pm$ 0.38\\
     & MMLU Humanities & A   & 59.36 $\pm$ 0.69\\
     & MMLU Other & A   & 71.58 $\pm$ 0.79\\
     & MMLU Social Sciences & A   & 74.39 $\pm$ 0.77\\
     & MMLU Stem & A   & 51.89 $\pm$ 0.84\\
     & TruthfulQA Gen & BA  & 45.78 $\pm$ 1.74\\
     & TruthfulQA Gen & BD  & -32.95 $\pm$ 31.31\\
     & TruthfulQA MC1 & A   & 39.53 $\pm$ 1.71\\
     & TruthfulQA MC2 & A   & 54.62 $\pm$ 1.59\\
\midrule

Q5\_K\_S & GSM8K (FE) & EM  & 75.66 $\pm$ 1.18\\
      & GSM8K (SM) & EM  & 17.13 $\pm$ 1.04\\
      & HellaSwag & A   & 57.19 $\pm$ 0.49\\
      & HellaSwag & AN  & 72.67 $\pm$ 0.44\\
      & IFEval & ILL & 84.53\\
      & IFEval & ILS & 81.53\\
      & IFEval & PLL & 77.82 $\pm$ 1.79\\
      & IFEval & PLS & 74.12 $\pm$ 1.88\\
      & MMLU & A   & 63.36 $\pm$ 0.38\\
      & MMLU Humanities & A   & 59.34 $\pm$ 0.69\\
      & MMLU Other & A   & 71.81 $\pm$ 0.78\\
      & MMLU Social Sciences & A   & 73.77 $\pm$ 0.77\\
      & MMLU Stem & A   & 50.87 $\pm$ 0.84\\
      & TruthfulQA Gen & BA  & 45.65 $\pm$ 1.74\\
      & TruthfulQA Gen & BD  & -13.25 $\pm$ 23.12\\
      & TruthfulQA MC1 & A   & 38.92 $\pm$ 1.71\\
      & TruthfulQA MC2 & A   & 53.90 $\pm$ 1.59\\
\midrule

Q5\_K\_M & GSM8K (FE) & EM  & 78.54 $\pm$ 1.13\\
      & GSM8K (SM) & EM  & 19.41 $\pm$ 1.09\\
      & HellaSwag & A   & 57.16 $\pm$ 0.49\\
      & HellaSwag & AN  & 72.33 $\pm$ 0.45\\
      & IFEval & ILL & 83.69\\
      & IFEval & ILS & 80.70\\
      & IFEval & PLL & 76.89 $\pm$ 1.81\\
      & IFEval & PLS & 73.38 $\pm$ 1.90\\
      & MMLU & A   & 62.80 $\pm$ 0.39\\
      & MMLU Humanities & A   & 58.85 $\pm$ 0.69\\
      & MMLU Other & A   & 71.39 $\pm$ 0.79\\
      & MMLU Social Sciences & A   & 73.68 $\pm$ 0.78\\
      & MMLU Stem & A   & 49.64 $\pm$ 0.84\\
      & TruthfulQA Gen & BA  & 43.82 $\pm$ 1.74\\
      & TruthfulQA Gen & BD  & 2.38 $\pm$ 22.63\\
      & TruthfulQA MC1 & A   & 38.56 $\pm$ 1.70\\
      & TruthfulQA MC2 & A   & 54.45 $\pm$ 1.59\\
\midrule
\midrule

Q6\_K & GSM8K (FE) & EM  & 78.17 $\pm$ 1.14\\
     & GSM8K (SM) & EM  & 20.17 $\pm$ 1.11\\
     & HellaSwag & A   & 57.37 $\pm$ 0.49\\
     & HellaSwag & AN  & 72.48 $\pm$ 0.45\\
     & IFEval & ILL & 82.85\\
     & IFEval & ILS & 79.98\\
     & IFEval & PLL & 75.60 $\pm$ 1.85\\
     & IFEval & PLS & 72.09 $\pm$ 1.93\\
     & MMLU & A   & 63.17 $\pm$ 0.38\\
     & MMLU Humanities & A   & 59.09 $\pm$ 0.69\\
     & MMLU Other & A   & 71.29 $\pm$ 0.79\\
     & MMLU Social Sciences & A   & 74.26 $\pm$ 0.77\\
     & MMLU Stem & A   & 50.46 $\pm$ 0.85\\
     & TruthfulQA Gen & BA  & 47.86 $\pm$ 1.75\\
     & TruthfulQA Gen & BD  & -4.51 $\pm$ 25.13\\
     & TruthfulQA MC1 & A   & 39.41 $\pm$ 1.71\\
     & TruthfulQA MC2 & A   & 54.71 $\pm$ 1.59\\
\midrule
\midrule

Q8\_0 & GSM8K (FE) & EM  & 77.48 $\pm$ 1.15\\
     & GSM8K (SM) & EM  & 23.50 $\pm$ 1.17\\
     & HellaSwag & A   & 57.42 $\pm$ 0.49\\
     & HellaSwag & AN  & 72.52 $\pm$ 0.45\\
     & IFEval & ILL & 83.81\\
     & IFEval & ILS & 80.70\\
     & IFEval & PLL & 77.45 $\pm$ 1.80\\
     & IFEval & PLS & 73.20 $\pm$ 1.91\\
     & MMLU & A   & 63.43 $\pm$ 0.38\\
     & MMLU Humanities & A   & 59.13 $\pm$ 0.68\\
     & MMLU Other & A   & 71.90 $\pm$ 0.78\\
     & MMLU Social Sciences & A   & 74.59 $\pm$ 0.76\\
     & MMLU Stem & A   & 50.62 $\pm$ 0.84\\
     & TruthfulQA Gen & BA  & 46.02 $\pm$ 1.74\\
     & TruthfulQA Gen & BD  & -16.41 $\pm$ 25.01\\
     & TruthfulQA MC1 & A   & 39.41 $\pm$ 1.71\\
     & TruthfulQA MC2 & A   & 54.81 $\pm$ 1.60\\

\bottomrule
\caption{Scores are reported in percentage points (mean $\pm$ standard error), with both the mean and standard error multiplied by 100.}
\label{tab:benchmark-quant}
\end{longtable}

\end{document}